\documentclass[runningheads]{llncs}

\usepackage[colorlinks,linkcolor=red,anchorcolor=blue, citecolor=blue]{hyperref}
\usepackage{multirow}
\usepackage[T1]{fontenc}
\usepackage{graphicx}
\usepackage{cite}
\usepackage{subcaption}
\usepackage{tabularx}
\usepackage{float}
\usepackage{amsmath} 
\usepackage{amssymb}
\usepackage{booktabs}
\usepackage{todonotes}
\usepackage{color}
\usepackage{overpic}
\usepackage{bm}
\usepackage{bbding}
\usepackage{makecell}
\usepackage{array}
\usepackage{caption}
\usepackage{diagbox}

\usepackage[T1]{fontenc}
\usepackage{graphicx}
\begin{document}
\title{A Multimodal Clinically Informed Coarse-to-Fine Framework for Longitudinal CT Registration in Proton Therapy}
\titlerunning{A Multimodal Clinically Informed Coarse-to-Fine Registration Framework}
% \author{***}
% \institute{}

\author{Caiwen Jiang \inst{1} \and
Yuzhen Ding \inst{1} \and
Mi Jia \inst{2} \and
Samir H. Patel \inst{1} \and
Terence T. Sio \inst{1} \and
Jonathan B. Ashman \inst{1} \and
Lisa A. McGee \inst{1} \and
Jean-Claude M. Rwigema \inst{1} \and
William G. Rule \inst{1} \and
Sameer R. Keole \inst{1} \and
Sujay A. Vora \inst{1} \and
William W. Wong \inst{1} \and
Nathan Y. Yu \inst{1} \and
Michele Y. Halyard \inst{1} \and
Steven E. Schild \inst{1} \and
Dinggang Shen \inst{2,3,4,5} \and
Wei Liu \inst{1}}
\institute{
Department of Radiation Oncology, Mayo Clinic, Phoenix, Arizona, USA\\
\email{jiang.caiwen@mayo.edu}
\and
School of Biomedical Engineering, ShanghaiTech University, Shanghai, China
\and
Shanghai United Imaging Intelligence Co., Ltd., Shanghai 200230, China
\and
Shanghai Artificial Intelligence Laboratory, Shanghai 200232, China
\and
Shanghai Clinical Research and Trial Center, Shanghai 201210, China
}
\maketitle              % typeset the header of the contribution
\vspace{-3mm}
\begin{abstract}
Proton therapy offers superior organ-at-risk sparing but is highly sensitive to anatomical changes, making accurate deformable image registration (DIR) across longitudinal CT scans essential. Conventional DIR methods are often too slow for emerging online adaptive workflows, while existing deep learning–based approaches are primarily designed for generic benchmarks and underutilize clinically relevant information beyond images. To this end, we propose a clinically scalable coarse-to-fine deformable registration framework that integrates multimodal information from the proton radiotherapy workflow to accommodate diverse clinical scenarios. The model employs dual CNN-based encoders for hierarchical feature extraction and a transformer-based decoder to progressively refine deformation fields. Beyond CT intensities, clinically critical priors, including target and organ-at-risk contours, dose distributions, and treatment planning text, are incorporated through \textbf{anatomy- and risk-guided attention}, \textbf{text-conditioned feature modulation}, and \textbf{foreground-aware optimization}, enabling anatomically focused and clinically informed deformation estimation. We evaluate the proposed framework on a large-scale and, to our knowledge, among the largest and most comprehensive proton therapy DIR datasets reported to date, comprising 1,222 paired planning and repeat CT scans across multiple anatomical regions and disease types. Extensive experiments demonstrate consistent improvements over state-of-the-art methods, enabling fast and robust clinically meaningful registration.

\vspace{-2mm}
\end{abstract}
\vspace{-3mm}
\vspace{-2mm}
\section{Introduction}
\vspace{-2mm}
% Compared with conventional radiotherapy, proton therapy enables radiation dose to be delivered more precisely to tumors while minimizing exposure to surrounding healthy tissues, making it particularly advantageous for treating tumors in proximity to critical organs~\cite{ding2025ai}. However, this benefit comes at the cost of increased sensitivity to accurate anatomical representation during treatment, and changes in patient positioning, organ motion, and treatment-induced deformation can result in clinically meaningful dose deviations~\cite{vestergaard2023range}. To mitigate these effects, adaptive proton therapy updates treatment plans using longitudinal CT scans acquired throughout the treatment course, which requires accurate alignment between planning and repeat CT images~\cite{palaniappan2021prdr}. In this context, deformable image registration (DIR) plays a central role by establishing voxel-level correspondence across time points and is therefore essential for safe and effective adaptive proton therapy.

Compared with conventional radiotherapy, proton therapy enables precise tumor irradiation with superior organ-at-risk sparing but is highly sensitive to accurate anatomical representation during treatment, where positioning changes, organ motion, and deformation can cause clinically significant dose deviations~\cite{vestergaard2023range}. Adaptive proton therapy mitigates this issue by updating treatment plans using longitudinal CT scans acquired throughout the treatment course, which requires accurate alignment between planning and repeat CT images~\cite{palaniappan2021prdr}. In this context, deformable image registration (DIR) is essential for establishing voxel-level correspondence across time points\cite{ding2025ai}. However, conventional DIR methods rely on computationally intensive iterative optimization, leading to long runtimes and limited scalability for time-critical online adaptive workflows.

The limitations of the conventional DIR have driven the rapid development of deep learning–based deformable registration. VoxelMorph introduced a seminal framework for directly predicting dense deformation fields via image-similarity optimization~\cite{balakrishnan2019voxelmorph}, later extended by multi-resolution coarse-to-fine architectures such as Dual-PRNet~\cite{kang2022dual} and CorrMLP~\cite{meng2024corrmlp}, as well as transformer-based models including TransMorph~\cite{chen2022transmorph}, to better capture large deformations and long-range correspondences. Despite strong benchmark performance, these approaches remain poorly aligned with radiotherapy workflows characterized by complex inter-fractional patient setup variations and anatomical changes, imposing substantially stricter accuracy and robustness requirements, particularly in proton therapy.

 \begin{figure}[!t]
 \setlength{\abovecaptionskip}{0.1cm}
\setlength{\belowcaptionskip}{-0.4cm}
\centering
\begin{overpic}[width=1\linewidth]{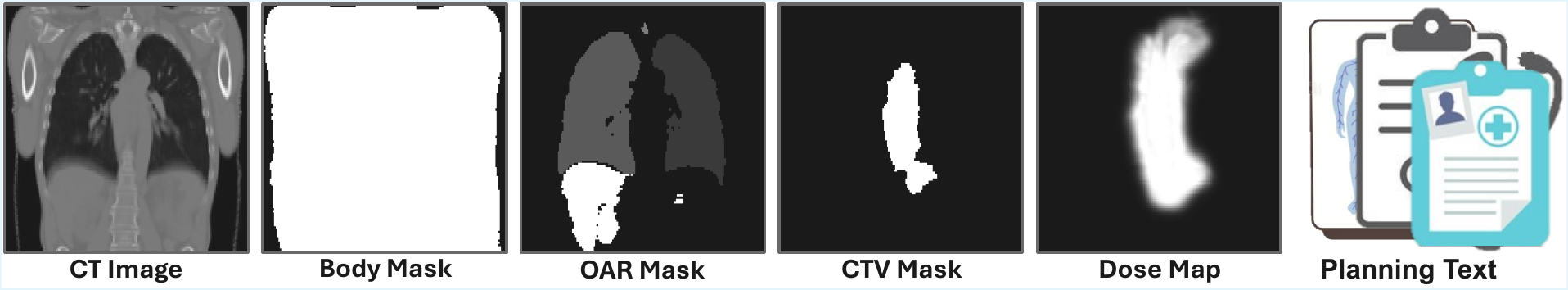}
    % \put(4, 47.5){\footnotesize (a) LPET}
        
    %     \put(13.7, 47.5){\footnotesize (b) Cycle-GAN}
    %     \put(27, 47.5){\footnotesize (c) 3D-cGAN}
    %     \put(39, 47.5){\footnotesize (d) CT-assisted}
    %     \put(52, 47.5){\footnotesize (e) LA-GAN}
    %     \put(63.5, 47.5){\footnotesize (f) AR-GAN}
    %     \put(77, 47.5){\footnotesize (g) Ours}
    %     \put(90, 47.5){\footnotesize (h) GT}
    \end{overpic}
    \vspace{-4mm}
\centering

\caption{Multimodal information available in the proton radiotherapy workflow for each CT scan, including body mask, OAR and CTV contours, dose distribution, and treatment planning texts (diagnosis and treatment planning notes).}
\vspace{-4mm}
\label{examples}
\end{figure}

Accordingly, several studies have attempted to incorporate radiotherapy workflow priors to adapt general-purpose DIR models for clinical applications. Representative efforts include contour-guided registration that enforces anatomical consistency within targets and organs-at-risk~\cite{gu2013cgdir, hemon2023contour}, as well as continuous spatiotemporal modeling frameworks that explicitly account for anatomical changes during the treatment course to better handle large deformations and sliding boundaries~\cite{li2024cptdir}.   However, these approaches are typically developed under idealized single-site or single-disease settings and trained on small clinical cohorts, often comprising only tens of patients, yielding models tailored to narrowly defined conditions. Moreover, reliance on CT intensities alone neglects rich multimodal clinical information in the radiotherapy workflow, biasing optimization toward global alignment and limiting accuracy within small but clinically critical target structures.

To address these challenges, we propose a clinically scalable deformable registration framework for large-deformation longitudinal CT alignment in proton radiotherapy. The framework leverages multimodal information from the proton therapy workflow to achieve robust performance across heterogeneous clinical scenarios. Architecturally, it adopts a progressive coarse-to-fine design with dual CNN encoders for hierarchical feature extraction and a transformer-based decoder for incremental deformation refinement capturing both global context and local anatomical variation.  Beyond CT images, we explicitly incorporate clinically relevant multimodal priors, including clinical target volume (CTV) and organ-at-risk (OAR) contours, dose distributions, and treatment planning texts (Fig.~\ref{examples}). These priors are integrated through anatomy- and risk-guided attention, text-conditioned feature modulation, and foreground-aware optimization, guiding deformation estimation toward anatomically coherent and clinically meaningful alignment. Finally, we curate a large-scale clinical proton radiotherapy dataset comprising 1,222 paired planning and repeat CT scans from 553 patients across diverse anatomical regions and disease categories, enabling the proposed framework to generalize across heterogeneous real-world clinical scenarios.

The main contributions of our work include 1) the first systematic integration of multimodal proton radiotherapy information for DIR, 2) constructing a progressive coarse-to-fine framework together with, to our knowledge, one of the largest and most comprehensive proton radiotherapy DIR dataset, and 3) achieving performance surpassing state-of-the-art methods.

 \begin{figure}[!t]
 \setlength{\abovecaptionskip}{0.1cm}
\setlength{\belowcaptionskip}{-0.4cm}
\centering
\begin{overpic}[width=1\linewidth]{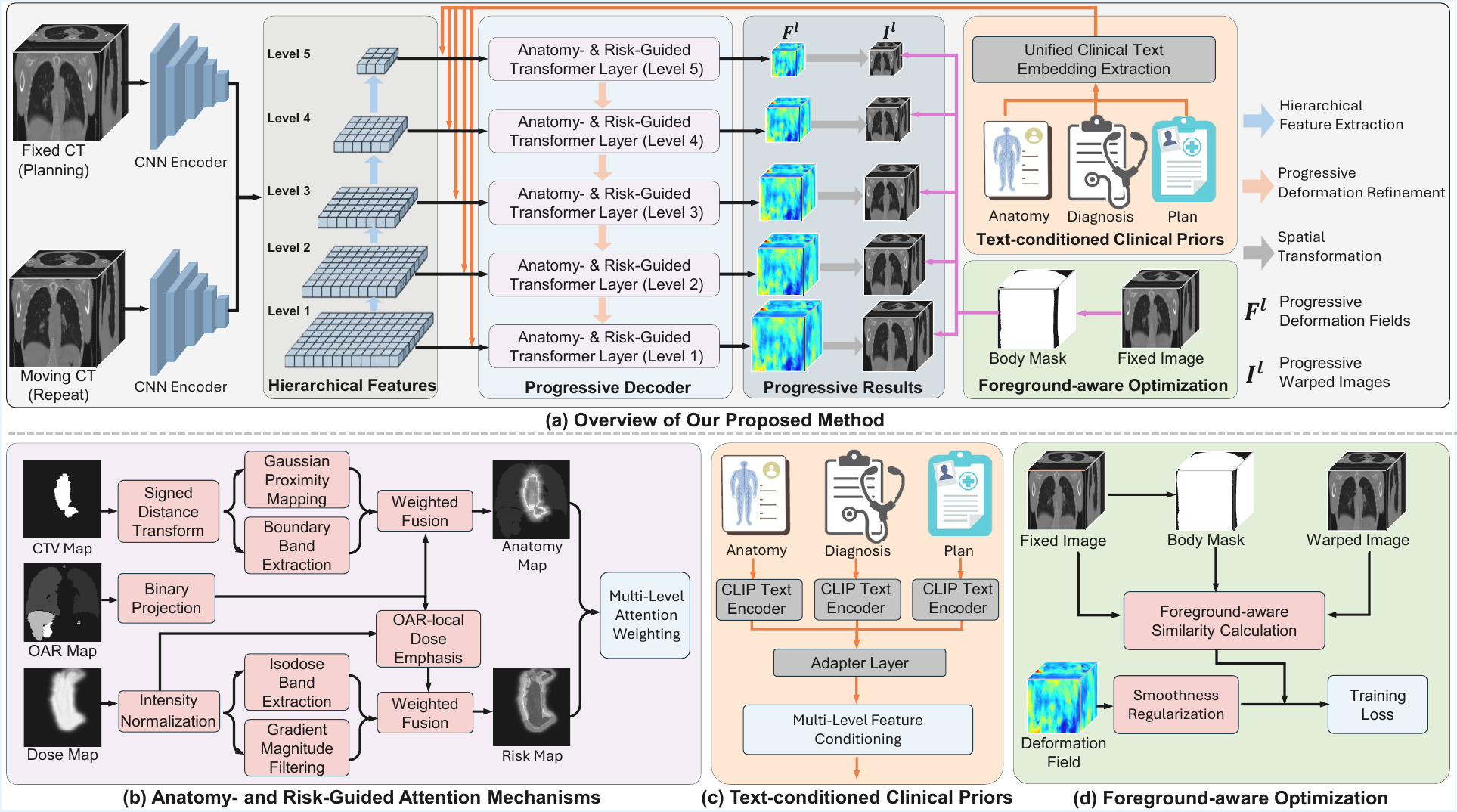}
    % \put(4, 47.5){\footnotesize (a) LPET}
        
    %     \put(13.7, 47.5){\footnotesize (b) Cycle-GAN}
    %     \put(27, 47.5){\footnotesize (c) 3D-cGAN}
    %     \put(39, 47.5){\footnotesize (d) CT-assisted}
    %     \put(52, 47.5){\footnotesize (e) LA-GAN}
    %     \put(63.5, 47.5){\footnotesize (f) AR-GAN}
    %     \put(77, 47.5){\footnotesize (g) Ours}
    %     \put(90, 47.5){\footnotesize (h) GT}
    \end{overpic}
    \vspace{-4mm}
\centering

\caption{Overview of the proposed multimodal clinically informed coarse-to-fine registration framework, highlighting anatomy- and risk-guided attention, text-conditioned clinical priors, and foreground-aware optimization.}
\vspace{-3mm}
\label{framework}
\end{figure}

\section{Method}
\vspace{-1.5mm}
\subsection{Progressive Coarse-to-Fine Transformer Framework for Large-Deformation Registration}
\vspace{-2mm}
Fig.~\ref{framework} illustrates the proposed progressive coarse-to-fine registration framework, designed to capture the large and complex deformations commonly observed in longitudinal radiotherapy CT scans due to anatomical variation and patient repositioning. The framework integrates weight-shared 3D CNN encoders for hierarchical feature extraction from the fixed and moving images with a progressive transformer-based decoder for coarse-to-fine deformation estimation. The CNN encoder comprises five convolutional stages, each consisting of two $3\times3\times3$ convolutional layers followed by instance normalization and LeakyReLU activation. Spatial resolution is progressively reduced by average pooling between stages, yielding multi-scale feature pyramids for both fixed and moving images. Corresponding feature levels are subsequently fused to form a five-level hierarchical representation $\{F^{l}\}_{l=1}^{5}$ for deformation refinement.

These fused multi-scale features are then processed by a Swin Transformer~\cite{liu2021swin} based decoder in a progressive coarse-to-fine manner. At each resolution level $l\in\{5,4,3,2,1\}$, the fused features are modeled by a transformer stage to capture long-range contextual dependencies and deformation cues, where anatomy- and risk-guided attention as well as text-conditioned feature modulation are integrated to provide clinically informed guidance, as detailed in Sec.~\ref{2.2} and~\ref{2.3}. The deformation field estimated at the previous coarser level is upsampled and used to warp the moving features toward the fixed space, providing an increasingly aligned initialization for subsequent refinement. A regression head predicts a residual displacement $\Delta\phi^{l}$, and the deformation is updated by additive composition,
$
\phi^{l} = \mathrm{Up}(\phi^{l+1}) + \Delta\phi^{l}.
$
This hierarchical refinement strategy enables stable modeling of large deformations at coarse scales while preserving fine anatomical structures at higher resolutions.

% The overview of our method is illustrated in Fig.~\ref{framework}(a). Given a pair of fixed and moving images, they are first processed by a shared 3D CNN encoder to extract hierarchical features, which are then fed into a progressive coarse-to-fine transformer-based decoder to iteratively refine deformation fields and produce the final output. Within this decoding process, \textbf{anatomy- and risk-guided attention} and \textbf{text-conditioned clinical priors} are integrated at multiple levels, with a \textbf{foreground-aware optimization} strategy is applied during training. The following sections detail these three components.
\vspace{-3.5mm}
\subsection{Anatomy- and Risk-Guided Attention for Anatomically Focused Alignment}\label{2.2}
\vspace{-2mm}

Although progressive transformer-based refinement effectively captures large deformations, appearance-driven similarity optimization is biased toward homogeneous regions, often compromising alignment accuracy within small but clinically critical targets and organs-at-risk~\cite{mok2020lapirn,zhao2019recursive}. We therefore introduce anatomy- and risk-guided attention to inject structural and treatment-related priors into multi-level deformation refinement, explicitly steering alignment toward anatomically focused and clinically meaningful regions. Specifically, anatomy and risk maps are constructed from clinical annotations and dose information,as illustrated in Fig.~\ref{framework}(b). The anatomy map integrates a signed distance–based Gaussian proximity map from the CTV, a boundary band emphasizing target interfaces, and OAR masks, which are fused into a continuous importance map. The risk map is derived from normalized dose distributions by combining dose gradients, high-dose isodose bands, and dose-weighted OAR regions, yielding a soft spatial prior that encodes treatment-related sensitivity to misalignment.

The anatomy and risk maps are fused into a unified prior representation and injected into the transformer decoder via multi-level attention weighting. At each decoding level, the fused prior is downsampled to match the feature resolution and projected into a gating signal through a lightweight convolution followed by sigmoid activation. This gating adaptively modulates feature responses, amplifying deformation updates in anatomically and clinically important regions while preserving global consistency elsewhere. By applying this guidance across multiple refinement stages, the proposed mechanism enables anatomically focused alignment from coarse deformation modeling to fine-scale correction.

\vspace{-3.5mm}
\subsection{Text-conditioned Feature Modulation for Semantically Informed Deformation Modeling}\label{2.3}
\vspace{-2mm}
While anatomy- and risk-guided attention emphasizes spatially important regions, deformable registration in radiotherapy is also influenced by high-level clinical knowledge that cannot be captured by voxel-wise appearance alone, including anatomical context, disease characteristics, and treatment planning information. To incorporate this complementary semantic guidance, we introduce text-conditioned feature modulation that encodes heterogeneous clinical cues into compact embeddings to inform deformation estimation. Specifically, three categories of prompts are extracted: anatomical descriptions of involved structures, diagnostic information characterizing tumor location, and treatment planning cues summarizing prescribed dose levels and critical organ considerations.

As shown in Fig.~\ref{framework}(c), these prompts are independently encoded using a CLIP-based text encoder (ViT-B/32) pretrained on large-scale vision–language pairs~\cite{radford2021learning} to obtain semantic embeddings. To adapt these general-purpose representations to the medical imaging domain, the embeddings are refined through a lightweight adapter layer and then injected into the progressive transformer decoder via multi-level feature modulation. At each decoding level $l$, the adapted embedding generates scale and shift parameters for feature-wise linear modulation (FiLM)~\cite{perez2018film}, applied as $
\mathbf{F}^{l\prime} = \mathbf{F}^l \odot (1 + \boldsymbol{\gamma}^l) + \boldsymbol{\beta}^l.$
This semantic modulation across multiple resolutions allows clinical priors to progressively guide deformation refinement from coarse alignment to fine-scale correction.

\vspace{-3.5mm}
\subsection{Foreground-Aware Optimization for Clinically Critical Deformation Learning}
\vspace{-2mm}

Similarity-based registration optimization is inherently dominated by background regions in volumetric images, reducing the impact of critical structures and posing particular risks in proton radiotherapy. Therefore, beyond anatomy- and risk-guided attention, we further introduce a foreground-aware optimization strategy to explicitly focus similarity learning on anatomically relevant regions. As illustrated in Fig.~\ref{framework}(d), foreground masks derived from the body contour of the fixed (planning) CT, which are routinely available in the radiotherapy workflow, are applied to both fixed and moving images during training to restrict similarity computation to anatomically relevant regions, thereby effectively suppressing background dominance. Unlike preprocessing-based cropping, our strategy operates directly at the loss level, where background regions do not contribute to optimization. Thus, zero-padding can be safely applied to standardize inputs, enabling consistent training across heterogeneous image sizes.

% The overall training objective combines a foreground-weighted image similarity term with a deformation smoothness regularizer. Specifically, masked normalized cross-correlation (NCC) is evaluated within foreground regions, while a regularization term penalizes excessive spatial gradients of the deformation field. The final loss is defined as
% \begin{equation}
% \mathcal{L}(\phi)
% =
% - \mathbb{E}_{x \sim p_{\mathrm{fg}}}
% \big[
% \mathrm{NCC}\big(I_f(x),\, I_m(\phi(x))\big)
% \big]
% +
% \lambda \mathcal{R}(\phi),
% \quad
% \mathcal{R}(\phi)=\|\nabla \phi\|_2^2 ,
% \label{train_loss}
% \end{equation}
% where $\phi$ denotes the deformation field, $I_f$ and $I_m$ are the fixed and moving images, and $\lambda$ balances alignment accuracy and smoothness.

The overall training objective combines a foreground-masked image similarity term with a deformation smoothness regularizer. Specifically, masked normalized cross-correlation (NCC) is evaluated within foreground regions, while a regularization term penalizes excessive spatial gradients of the deformation field. The final loss is defined as
\begin{equation}
\mathcal{L}(\phi)=
-\mathrm{NCC}\!\left(M_{\mathrm{fg}}\odot I_f,\; M_{\mathrm{fg}}\odot (I_m\circ\phi)\right)
+\lambda\|\nabla\phi\|_2^2 ,
\label{train_loss}
\end{equation}
where $M_{\mathrm{fg}}$ denotes the foreground mask, $\odot$ indicates element-wise multiplication, $I_f$ and $I_m$ are the fixed and moving images, $\phi$ is the deformation field, and $\lambda$ balances alignment accuracy and smoothness.

\section{Experiments}
\vspace{-2mm}
\subsection{Dataset and Implementation Details}
\begin{figure}[!t]
\setlength{\abovecaptionskip}{0.1cm}
\setlength{\belowcaptionskip}{-0.4cm}
\centering

\newcommand{\leftH}{0.25\textheight}
\newcommand{\rightH}{0.105\textheight}

\begin{minipage}[t]{0.6\linewidth}
    \centering
    \includegraphics[height=\leftH, width=\linewidth]{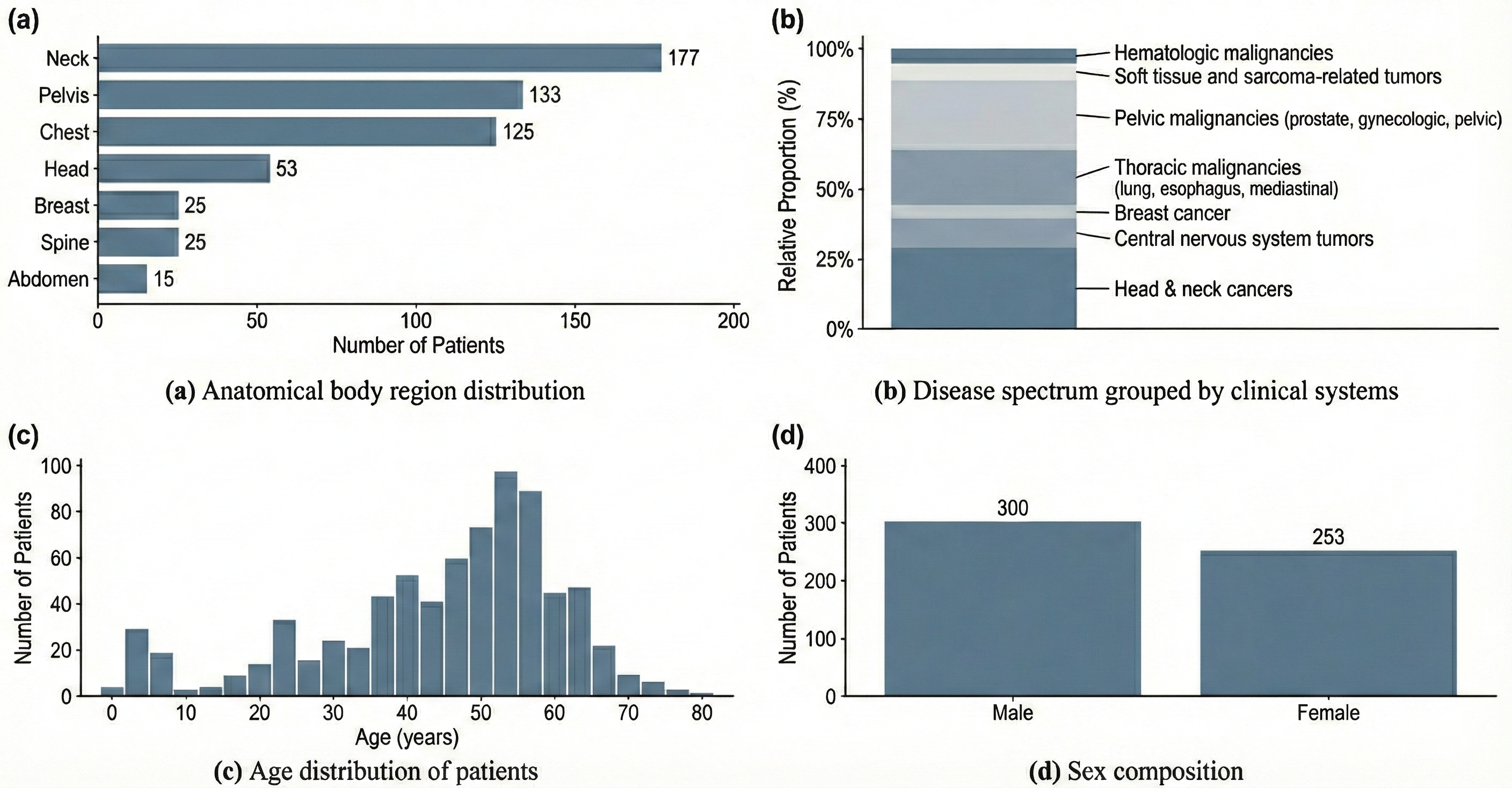}
\captionof{figure}{Cohort overview of the dataset.}\label{fig:dataset_overview}
\end{minipage}\hfill
\begin{minipage}[t]{0.4\linewidth}\vspace*{-48mm}
    \centering
    \includegraphics[height=\rightH, width=\linewidth]{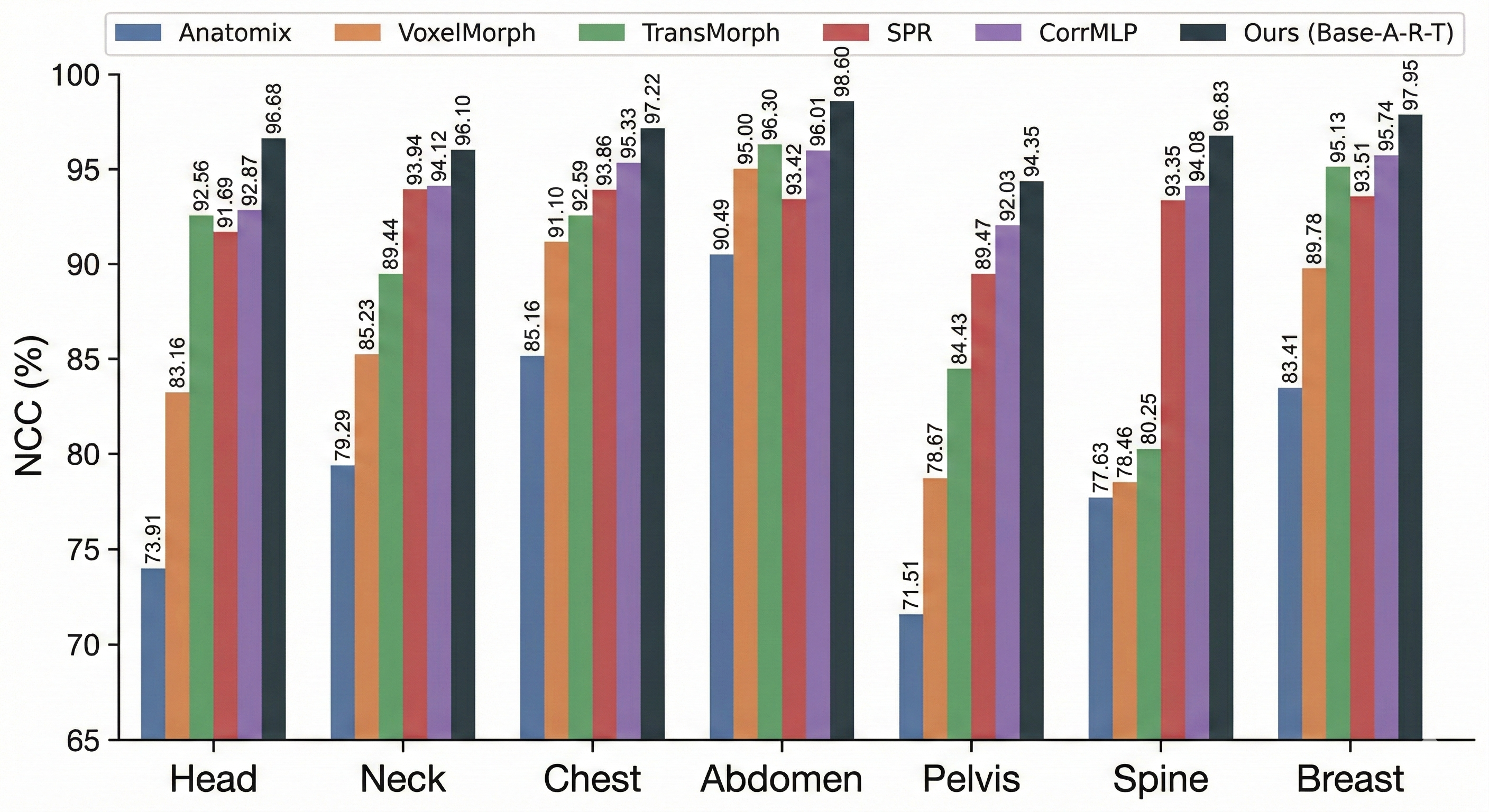}\vspace*{-0.5mm}
\captionof{figure}{Multi-region results.}\label{fig:multi_region_stats}
\vspace*{5mm}
    \includegraphics[height=\rightH, width=\linewidth]{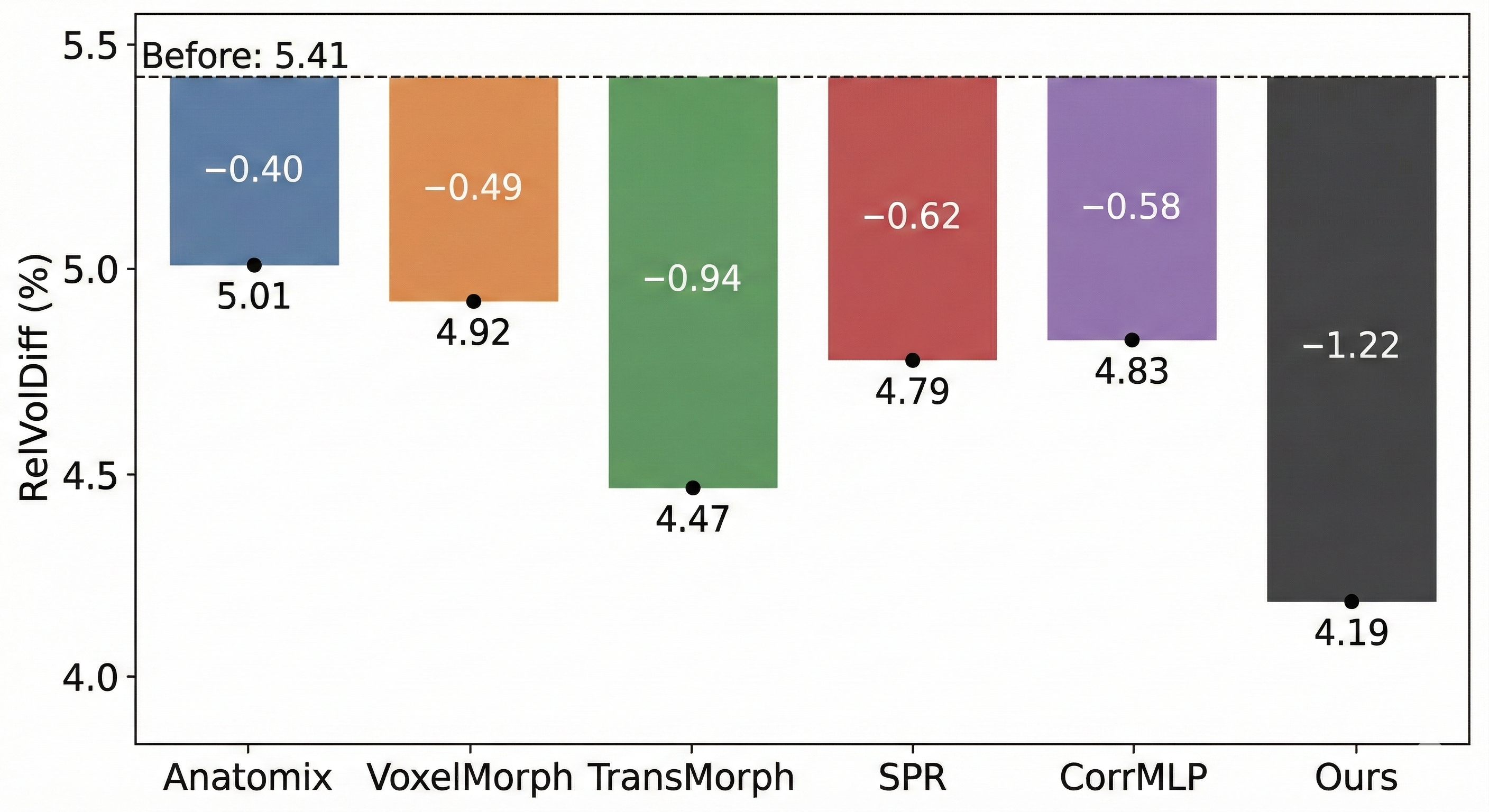}
\captionof{figure}{CTV propagation.}\label{fig:ctv_results}
\end{minipage}
\vspace{-2mm}
% \caption{Clinical Cohort Overview of Proton Radiotherapy CT Dataset.}
% \vspace{-4mm}
\label{dataset_multi_ctv}
\end{figure}

\vspace{-2mm}
We evaluate the proposed framework on a large-scale clinical proton radiotherapy dataset collected from routine clinical practice, comprising longitudinal CT scans from 553 patients with a total of 1,338 volumetric images. By randomly pairing CT scans from the same patient across sessions, we construct 1,222 fixed–moving image pairs for deformable registration. The dataset spans diverse anatomical sites, disease types, age ranges, and patient demographics, with detailed distributions illustrated in Fig.~\ref{fig:dataset_overview}.

For all anatomical regions, 80\% of the image pairs are used for training and the remaining 20\% for testing, with patient-level separation to prevent data leakage. All volumes are zero-padded to a unified spatial resolution of $512 \times 512 \times 256$ to accommodate varying image sizes. Each image pair is further pre-aligned into a common coordinate space using rigid registration to provide consistent initialization for deformable alignment. The network is trained on four NVIDIA A100 GPUs using data-parallel optimization with a polynomial learning rate decay schedule and an initial learning rate of $1\times10^{-4}$. The regularization weight $\lambda$ in Eq.~\ref{train_loss} is empirically set to $0.2$ across all experiments.

%  \begin{figure}[!t]
%  \setlength{\abovecaptionskip}{0.1cm}
% \setlength{\belowcaptionskip}{-0.4cm}
% \centering
% \begin{overpic}[width=1\linewidth]{dataset-crop.PNG}
%     % \put(4, 47.5){\footnotesize (a) LPET}
        
%     %     \put(13.7, 47.5){\footnotesize (b) Cycle-GAN}
%     %     \put(27, 47.5){\footnotesize (c) 3D-cGAN}
%     %     \put(39, 47.5){\footnotesize (d) CT-assisted}
%     %     \put(52, 47.5){\footnotesize (e) LA-GAN}
%     %     \put(63.5, 47.5){\footnotesize (f) AR-GAN}
%     %     \put(77, 47.5){\footnotesize (g) Ours}
%     %     \put(90, 47.5){\footnotesize (h) GT}
%     \end{overpic}
%     \vspace{-4mm}
% \centering

% \caption{Clinical Cohort Overview of Proton Radiotherapy CT Dataset.}
% \vspace{-4mm}
% \label{dataset}
% \end{figure}

% \begin{figure}[!t]
% \setlength{\abovecaptionskip}{0.1cm}
% \setlength{\belowcaptionskip}{-0.4cm}
% \centering
% \includegraphics[width=0.8\linewidth]{dataset-crop.PNG}
% \caption{Cohort overview of the dataset.}
% \label{fig:dataset_overview}
% \vspace{-1mm}
% \end{figure}

% \begin{figure}[!t]
% \setlength{\abovecaptionskip}{0.1cm}
% \setlength{\belowcaptionskip}{-0.4cm}
% \centering
% \includegraphics[width=0.7\linewidth]{Multi_region.png}
% \caption{Multi-region results.}
% \label{fig:multi_region_stats}
% \vspace{-3mm}
% \end{figure}

% \begin{figure}[!t]
% \setlength{\abovecaptionskip}{0.1cm}
% \setlength{\belowcaptionskip}{-0.4cm}
% \centering
% \includegraphics[width=0.7\linewidth]{CTV_result.png}
% \caption{CTV propagation.}
% \label{fig:ctv_results}
% \vspace{-3mm}
% \end{figure}

\vspace{-3.5mm}
\subsection{Component-wise Ablation Analysis}
\vspace{-2mm}
To evaluate the contribution of each component, we design five ablation variants, namely Base, Base-A, Base-R, Base-A-R, and Base-A-R-T, as summarized in Table~\ref{tab:comparison-ablation}. Base denotes the progressive coarse-to-fine transformer baseline without clinical priors. Base-A and Base-R respectively introduce anatomy-guided and risk-guided attention, Base-A-R combines both spatial priors, and Base-A-R-T further incorporates text-conditioned clinical priors.

Several important observations can be drawn from the results. First, introducing either anatomy-guided or risk-guided attention consistently improves performance over the baseline, with NCC increasing from 95.43\% (Base) to 96.64\% (Base-A) and 96.57\% (Base-R), respectively, demonstrating the benefit of explicitly emphasizing clinically relevant regions. Second, combining both forms of spatial guidance further enhances alignment accuracy, yielding a NCC  of 96.72\% and improved SSIM, which indicates their complementary roles in capturing structural importance and treatment-related sensitivity. Third, incorporating text-conditioned clinical priors achieves the best overall performance, reaching a NCC of 96.82\% and a SSIM of 89.13\% with the lowest variance, highlighting the added value of integrating high-level clinical knowledge beyond voxel-wise appearance. Overall, these results confirm that each component contributes positively to the deformation accuracy, with their joint integration providing the most robust and precise registration.

\begin{table}[!t]
\centering
\setlength{\tabcolsep}{3.5pt}
\renewcommand{\arraystretch}{1.15}
% Make numbers visually larger while still fitting the page: scale slightly less than \textwidth,
% and apply \Large inside the resized box (so it affects the scaled content).
\resizebox{0.98\textwidth}{!}{%
\Large
\renewcommand{\scriptsize}{\footnotesize}%
\begin{tabular}{l|ccccc|ccccc}
\hline\hline
\multirow{2}{*}{\textbf{Metric}} & \multicolumn{5}{c|}{\textbf{Comparison}} & \multicolumn{5}{c}{\textbf{Ablation}} \\ \cline{2-11} 
 & Anatomix & VoxelMorph & TransMorph & SPR & CorrMLP & Base & Base-A & Base-R & Base-A-R & Base-A-R-T \\ 
\hline

NCC(\%) $\uparrow$ 
& 80.2$_{\scriptsize(10.64)}$ 
& 85.92$_{\scriptsize(5.16)}$ 
& 90.1$_{\scriptsize(6.24)}$ 
& 93.67$_{\scriptsize(5.76)}$ 
& \textbf{94.71$_{\scriptsize(4.41)}$} 
& 95.43$_{\scriptsize(3.61)}$ 
& 96.64$_{\scriptsize(2.25)}$ 
& 96.57$_{\scriptsize(2.56)}$ 
& 96.72$_{\scriptsize(1.96)}$ 
& \textbf{96.82$_{\scriptsize(1.69)}$} \\

MSE (\%) $\downarrow$ 
& 0.89$_{\scriptsize(0.0400)}$ 
& 0.53$_{\scriptsize(0.0196)}$ 
& 0.30$_{\scriptsize(0.0100)}$ 
& \textbf{0.11$_{\scriptsize(0.0036)}$} 
& \textbf{0.11$_{\scriptsize(0.0036)}$} 
& 0.13$_{\scriptsize(0.0049)}$ 
& 0.12$_{\scriptsize(0.0036)}$ 
& \textbf{0.11$_{\scriptsize(0.0036)}$} 
& \textbf{0.11$_{\scriptsize(0.0025)}$} 
& \textbf{0.11$_{\scriptsize(0.0025)}$} \\

SSIM(\%) $\uparrow$ 
& 71.0$_{\scriptsize(9.32)}$ 
& 79.7$_{\scriptsize(6.25)}$ 
& 78.99$_{\scriptsize(8.04)}$ 
& 85.89$_{\scriptsize(5.31)}$ 
& \textbf{86.79$_{\scriptsize(4.29)}$} 
& 88.34$_{\scriptsize(5.76)}$ 
& 88.74$_{\scriptsize(4.84)}$ 
& 88.55$_{\scriptsize(5.29)}$ 
& 89.06$_{\scriptsize(4.00)}$ 
& \textbf{89.13$_{\scriptsize(3.61)}$} \\

\hline
\Xhline{2\arrayrulewidth}
\end{tabular}
}
\caption{Quantitative comparison and ablation results across NCC, MSE, and SSIM.}
\label{tab:comparison-ablation}
\vspace{-9mm}
\end{table}

\vspace{-3.5mm}
\subsection{Comparison with State-of-the-Art Methods}
\vspace{-2mm}
We compare the proposed framework with five representative deformable registration paradigms, including feature-matching-based Anatomix~\cite{dey2024learning}, CNN-based VoxelMorph~\cite{balakrishnan2019voxelmorph}, transformer-based TransMorph~\cite{chen2022transmorph}, spatially adaptive regularization method SPR~\cite{chen2025unsupervised}, and the coarse-to-fine architecture CorrMLP~\cite{meng2024corrmlp}. These methods collectively cover major architectural and modeling strategies in contemporary deformable registration. All approaches are evaluated under the same experimental protocol for fair comparison.

\begin{figure}[!t]
 \setlength{\abovecaptionskip}{0.1cm}
\setlength{\belowcaptionskip}{-0.4cm}
\centering
\begin{overpic}[width=1\linewidth]{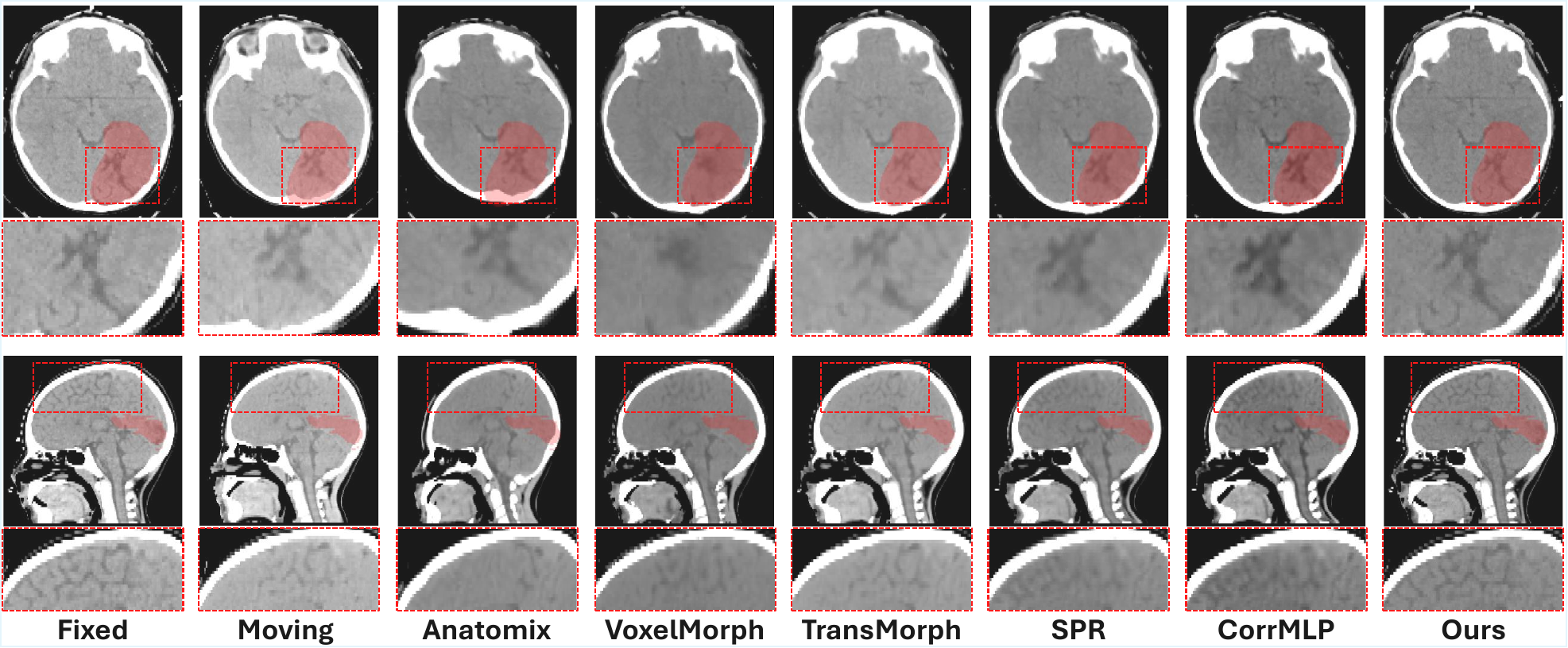}
    % \put(4, 47.5){\footnotesize (a) LPET}
        
    %     \put(13.7, 47.5){\footnotesize (b) Cycle-GAN}
    %     \put(27, 47.5){\footnotesize (c) 3D-cGAN}
    %     \put(39, 47.5){\footnotesize (d) CT-assisted}
    %     \put(52, 47.5){\footnotesize (e) LA-GAN}
    %     \put(63.5, 47.5){\footnotesize (f) AR-GAN}
    %     \put(77, 47.5){\footnotesize (g) Ours}
    %     \put(90, 47.5){\footnotesize (h) GT}
    \end{overpic}
    \vspace{-4mm}
\centering

\caption{Qualitative comparison of registration results for a representative head case across axial and sagittal views, with the CTV highlighted in red and zoomed regions showing local deformation differences.}
\vspace{-3mm}
\label{visual-result}
\end{figure}

Quantitative results in Table~\ref{tab:comparison-ablation} demonstrate that our method achieves the highest NCC (96.82\%) and SSIM (89.13\%) with consistently lower variance than competing approaches, indicating improved overall alignment accuracy and robustness. While recent methods such as SPR and CorrMLP outperform earlier CNN- and transformer-based methods, our framework further enhances performance by explicitly incorporating clinically informed spatial and semantic priors. This numerical advantage is clearly reflected in the qualitative comparisons in Fig.~\ref{visual-result}, where competing methods exhibit boundary blurring and local misalignment within the highlighted CTV regions, particularly under large cortical deformations in both axial and sagittal views. In contrast, our approach preserves sharper anatomical boundaries and achieves more coherent local alignment in the zoomed regions, closely matching the fixed anatomy, thereby confirming the effectiveness of clinically guided deformation refinement.

\vspace{-3.5mm}
\subsection{Generalization Performance across Anatomical Regions}
\vspace{-2mm}
To further evaluate the generalization ability of the proposed method under diverse clinical variations, we present results along one representative clinical factor, namely across multiple anatomical regions, including the head, neck, chest, abdomen, pelvis, spine, and breast, as summarized in Fig.~\ref{fig:multi_region_stats}. Our method consistently achieves the highest NCC across all evaluated regions, demonstrating strong generalization under anatomical variability. In contrast, competing methods exhibit larger performance fluctuations across regions, particularly in anatomically complex areas such as the pelvis and spine, while our framework maintains stable and high alignment accuracy. This consistent performance highlights the superior generalization of our clinically informed framework, enabled by progressive deformation modeling and explicit anatomical guidance that mitigate sensitivity to regional heterogeneity.

\vspace{-3.5mm}
\subsection{Clinical Utility in Target Propagation}
\vspace{-2mm}
Finally, we assess the clinical utility of the proposed framework in a key radiotherapy workflow task of propagating target contours from the planning CT (fixed) to the repeat CT (moving) via deformable registration to reduce manual re-delineation effort. Target alignment is quantified using the relative volume difference (RelVolDiff) between CTV regions before and after registration, where lower values indicate better correspondence. As shown in Fig.~\ref{fig:ctv_results}, deformable registration markedly reduces RelVolDiff from the unregistered baseline of 5.41\% for all methods. Our framework achieves the lowest value of 4.19\%, outperforming TransMorph (4.47\%) and other baselines (4.79–5.01\%), indicating more accurate preservation of small but clinically critical target structures. These results confirm that the proposed clinically guided framework translates improved alignment into tangible benefits for target propagation in adaptive proton therapy.

% \begin{table}[h]
% \centering
% \setlength{\tabcolsep}{3pt}
% \renewcommand{\arraystretch}{1.05}
% \resizebox{\textwidth}{!}{%
% % 定义列格式：第一列左对齐，中间5列居中，右边5列居中，中间用竖线分隔
% \begin{tabular}{l|ccccc|ccccc}
% \hline\hline
% \multirow{2}{*}{\textbf{Metric}} & \multicolumn{5}{c|}{\textbf{Comparison}} & \multicolumn{5}{c}{\textbf{Ablation}} \\ \cline{2-11} 
%  & Anatomix & VoxelMorph & TransMorph & SPR & CorrMLP & Base & Base-A & Base-R & Base-A-R & Base-A-R-T \\ \hline
% NCC(\%) $\uparrow$ & 80.2 & 85.92 & 90.1 & 93.67 & \textbf{94.71} & 95.43 & 96.64 & 96.57 & 96.72 & \textbf{96.82} \\
% MSE (\%) $\downarrow$ & 0.89 & 0.53 & 0.3 & \textbf{0.11} & \textbf{0.11} & 0.13 & 0.12 & \textbf{0.11} & \textbf{0.11} & \textbf{0.11} \\
% SSIM(\%) $\uparrow$ & 71 & 79.7 & 78.99 & 85.89 & \textbf{86.79} & 88.34 & 88.74 & 88.55 & 89.06 & \textbf{89.13} \\ \hline
% \Xhline{2\arrayrulewidth}
% \end{tabular}
% }%
% \caption{Quantitative comparison and ablation results.}
% \label{tab:comparison-ablation}
% \end{table}

\vspace{-4mm}
\section{Conclusion}
\vspace{-3mm}
In this paper, we present a clinically informed deformable image registration framework for longitudinal CT alignment in radiotherapy by integrating multimodal clinical priors into a progressive transformer-based architecture. The proposed method incorporates anatomy- and risk-guided attention for spatially focused deformation refinement, text-conditioned feature
modulation for high-level semantic conditioning, and a foreground-aware optimization strategy to emphasize clinically relevant anatomy during training. Extensive experiments on a large-scale proton radiotherapy dataset demonstrate consistent improvements over the state-of-the-art methods across quantitative metrics, anatomical regions, and clinically meaningful target propagation tasks. These results highlight the value of embedding clinical knowledge into deformable registration and its potential for adaptive radiotherapy applications.

% {\small
% \bibliography{myref}

{\small
\begin{thebibliography}{10}
\providecommand{\url}[1]{\texttt{#1}}
\providecommand{\urlprefix}{URL }
\providecommand{\doi}[1]{https://doi.org/#1}

\bibitem{balakrishnan2019voxelmorph}
Balakrishnan, G., Zhao, A., Sabuncu, M.R., Guttag, J., Dalca, A.V.: Voxelmorph: A learning framework for deformable medical image registration. IEEE Transactions on Medical Imaging  \textbf{38}(8),  1788--1800 (2019)

\bibitem{chen2022transmorph}
Chen, J., Frey, E., He, Y., Segars, W., Li, Y., Du, Y.: Transmorph: Transformer for unsupervised medical image registration. Medical Image Analysis  \textbf{82},  102615 (2022)

\bibitem{chen2025unsupervised}
Chen, J., Wei, S., Liu, Y., Bian, Z., He, Y., Carass, A., Bai, H., Du, Y.: Unsupervised learning of spatially varying regularization for diffeomorphic image registration. Medical Image Analysis p. 103887 (2025)

\bibitem{dalca2019probabilistic}
Dalca, V., Balakrishnan, G., Guttag, J., Sabuncu, R.: Unsupervised learning of probabilistic diffeomorphic registration for images and surfaces. Medical Image Analysis  \textbf{57},  226--236 (2019)

\bibitem{dey2024learning}
Dey, N., Billot, B., Wong, E., Wang, J., Ren, M., Grant, E., Dalca, V., Golland, P.: Learning general-purpose biomedical volume representations using randomized synthesis. arXiv preprint arXiv:2411.02372  (2024)

\bibitem{ding2025ai}
Ding, Y., Feng, H., Bues, M., Fatyga, M., Liu, T., Whitaker, J., Lin, H., Lee, Y., Simone, B., Patel, H., Ma, J., Frank, J., Vora, A., Ashman, A., Liu, W.: Ai in proton therapy treatment planning: A review. arXiv preprint arXiv:2510.19213  (2025)

\bibitem{gu2013cgdir}
Gu, X., Dong, B., Wang, J., Yordy, J., Mell, L., Jia, X., Jiang, S.B.: A contour-guided deformable image registration algorithm for adaptive radiotherapy. arXiv preprint arXiv:1301.0970  (2013)

\bibitem{hemon2023contour}
Hemon, C., Rigaud, B., Barateau, A., Tilquin, F., Noblet, V., Sarrut, D., Meyer, P., Bert, J., De~Crevoisier, R., Simon, A.: Contour-guided deep learning based deformable image registration for dose monitoring during {CBCT}-guided radiotherapy of prostate cancer. Journal of Applied Clinical Medical Physics  \textbf{24}(8),  e13991 (2023)

\bibitem{kang2022dual}
Kang, M., Hu, X., Huang, W., Scott, R., Reyes, M.: Dual-stream pyramid registration network. Medical Image Analysis  \textbf{78},  102379 (2022)

\bibitem{li2024cptdir}
Li, X., Li, M., Lomax, A., Buhmann, J., Zhang, Y.: Continuous spatial-temporal deformable image registration ({CPT-DIR}) for motion modelling in radiotherapy: beyond classic voxel-based methods. arXiv preprint arXiv:2405.00430  (2024)

\bibitem{liu2021swin}
Liu, Z., Lin, Y., Cao, Y., Hu, H., Wei, Y., Zhang, Z., Lin, S., Guo, B.: Swin transformer: Hierarchical vision transformer using shifted windows. In: Proceedings of the IEEE/CVF International Conference on Computer Vision (ICCV). pp. 10012--10022 (2021)

\bibitem{meng2024corrmlp}
Meng, M., Feng, D., Bi, L., Kim, J.: Correlation-aware coarse-to-fine {MLP}s for deformable medical image registration. arXiv preprint arXiv:2406.00123  (2024)

\bibitem{mok2020lapirn}
Mok, W., Chung, S.: Large deformation diffeomorphic image registration with laplacian pyramid networks. In: Medical Image Computing and Computer Assisted Intervention. pp. 561--570 (2020)

\bibitem{palaniappan2021prdr}
Palaniappan, P., Meyer, S., Kamp, F., Belka, C., Riboldi, M., Parodi, K., Gianoli, C.: Deformable image registration of the treatment planning {CT} with proton radiographies in perspective of adaptive proton therapy. Physics in Medicine \& Biology  \textbf{66}(4),  045008 (2021)

\bibitem{perez2018film}
Perez, E., Strub, F., de~Vries, H., Dumoulin, V., Courville, A.: Film: Visual reasoning with a general conditioning layer. In: Proceedings of the AAAI Conference on Artificial Intelligence. vol.~32 (2018)

\bibitem{radford2021learning}
Radford, A., Kim, J., Hallacy, C., Ramesh, A., Goh, G., Agarwal, S., Sastry, G., Askell, A., Mishkin, P., Clark, J., et~al.: Learning transferable visual models from natural language supervision. Proceedings of the International Conference on Machine Learning  \textbf{139},  8748--8763 (2021)

\bibitem{vestergaard2023range}
Vestergaard, D., Muren, P., Elstr{\o}m, V., Johansen, G., Taasti, T.: Tissue-specific range uncertainty estimation in proton therapy. Physics and Imaging in Radiation Oncology  \textbf{26},  100441 (2023)

\bibitem{zhang2021cascaded}
Zhang, L., Zhou, L., Li, R., Wang, X., Han, B., Liao, H.: Cascaded feature warping network for unsupervised medical image registration. arXiv preprint arXiv:2103.08213  (2021)

\bibitem{zhao2019recursive}
Zhao, S., Dong, Y., Chang, C., Xu, Y.: Recursive cascaded networks for unsupervised medical image registration. arXiv preprint arXiv:1907.12353  (2019)

\end{thebibliography}

}
% }

\end{document}